\documentclass{article} 
\PassOptionsToPackage{dvipsnames}{xcolor}
\usepackage{iclr2025_conference,times}

\usepackage{url}

\usepackage{graphicx}
\usepackage{wrapfig}
\usepackage{placeins}
\usepackage{hyperref}
\usepackage[hyperpageref]{backref} 
\usepackage[aboveskip=2pt, belowskip=0pt]{caption}
\usepackage{xcolor}
\usepackage{pifont}
\usepackage{multirow}
\newcommand{\cmark}{\color{ForestGreen}\ding{51}}%
\newcommand{\xmark}{\color{BrickRed}\ding{55}}%

\title{Assessing SAM for tree crown instance segmentation from drone imagery}

\author{Mélisande Teng \\
Mila, Université de Montréal \\
\And
Arthur Ouaknine \\
Mila, McGill University, Rubisco AI\\
\And
Etienne Laliberté\\
Université de Montréal \\
\And
Yoshua Bengio\\
Mila, Université de Montréal \\
\And
David Rolnick \\
Mila, McGill University\\
\And
Hugo Larochelle \\
Mila \\
}

\iclrfinalcopy 
\begin{document}

\maketitle

\begin{abstract}
The potential of tree planting as a natural climate solution is often undermined by inadequate monitoring of tree planting projects. Current monitoring methods involve measuring trees by hand for each species, requiring extensive cost, time, and labour. 
Advances in drone remote sensing and computer vision offer great potential for mapping and characterizing trees from aerial imagery, and large pre-trained vision models, such as the Segment Anything Model (SAM), may be a particularly compelling choice given limited labeled data.
In this work, we compare SAM methods for the task of automatic tree crown instance segmentation in high resolution drone imagery of young tree plantations. We explore the potential of SAM for this task, and find that methods using SAM out-of-the-box do not outperform a custom Mask R-CNN, even with well-designed prompts, but that there is potential for methods which tune SAM further. We also show that predictions can be improved by adding Digital Surface Model (DSM) information as an input. 

\end{abstract}

\section{Introduction}


Afforestation, reforestation, and revegetation have great potential as cost-effective natural climate solutions \citep{busch2024cost}, and monitoring carbon stocks in tree plantations is key to evaluate the success of such efforts to mitigate climate change \citep{canadell2008managing, di2021ten}. While tree plantations are still largely monitored by conducting manual tree surveys \citep{verra_vm0047}, recent advances in remote sensing and deep learning open possibilities for automatically estimating above ground biomass through the task of tree crown segmentation. Indeed, carbon stored in a tree can be recovered with allometries using information about the crown surface area, the species, and the height of the tree \citep{ jucker2022tallo}. 
Popular deep learning methods, such as Mask R-CNN \citep{he2017mask} and RetinaNet \citep{lin2017focal}, have been extensively used in the context of vegetation monitoring, but they most often do not focus on identifying tree species \citep{ball2023detectree, weinstein2019individual}.
Despite the success of deep learning methods for tree mapping at scale using remote sensing imagery \citep{tucker2023sub, reiersen2022reforestree}, instance segmentation of tree crowns remains understudied, in large part because of the lack of annotated data at the individual tree level. However, in the context of tree plantations, estimation of carbon stocks at the individual tree level is particularly important, especially when studying sites with multiple tree species as different species have different allometries \citep{singh2011formulating, daba2019accuracy,mulatu2024species}. \\
In contexts in which task-specific data is not abundant, practitioners often resort to using pre-trained models on large datasets.  The Segment Anything Model (SAM), designed for generalized object segmentation, offers a method for automatically segmenting any object in an image~\citep{kirillov2023segment} in a zero-shot setting, or given user input prompts in the form of points, boxes, masks or text. SAM has been used out-of-the-box for a wide variety of applications, such as medical imaging \citep{cheng2023sam} or river water segmentation in remote sensing imagery \citep{moghimi2024comparative}. However, despite its zero-shot abilities, SAM has been found to perform poorly in certain segmentation tasks when used directly in its automatic mode \citep{chen2023sam}, and consequently, a number of methods have been developed to adapt SAM to specific tasks without requiring fine-tuning it fully \citep{osco2023segment, segmate2023}. \\
In particular, RSPrompter \citep{chen2024rsprompter} proposed to learn how to generate appropriate prompts for SAM in order to segment objects of interest in remote sensing imagery. Keeping the image encoder and mask decoder frozen, a learnable prompter taking as input the image embeddings from the image encoder is trained to produce task-relevant prompts for the mask decoder. \\
In this work, we conduct the first study of tree crown instance segmentation on the UAV Canadian (Quebec) Plantations dataset \citep{Lefebvre2024}. Our contributions are: preparing the dataset for this task, and assessing  the relevance of SAM in the context of tree crown instance segmentation in high resolution drone imagery. We compare SAM's automatic mode with using prompts obtained from other object detection or instance segmentation models, or with prompts learned through parameter-efficient tuning of SAM. We also explore incorporating Digital Surface Model (DSM)  information, which is available from RGB drone photogrammetry, and find that doing so improves performance on our task. Building on RSPrompter, we present DSMPrompter, which outperforms other methods on all metrics. Our study highlights limitations of SAM in its intended use as an out-of-the-box, user-friendly tool, and the need for task-specific tuning for it to be advantageous over simpler custom trained models.

\section{Dataset} 
We use RGB orthomosaics, photogrammetry digital surface models (DSMs), and tree crown delineation and species labels in plantation sites from the UAV Canadian (Quebec) Plantations dataset \citep{Lefebvre2024}. The imagery has a resolution of 5mm/pixel. The DSM captures the elevation of the top surface of an area (including any object on the ground), and is obtained without any additional cost to the imagery acquisition by processing overlapping pictures.\\ 
We excluded the Serpentin1 and Serpentin2 sites from our study, because they contain respectively only 25 and 39 annotated trees and kept 15 sites of interest. 
We considered tree species that had more than 20 trees across all sites, and grouped the remaining species into an ``Other'' category, resulting in a total of 9 classes. Species category codes, corresponding scientific names, and number of trees per species per site are detailed in Table \ref{specie_name_correspondence} in \ref{appendix:data}.  The provided annotations correspond to the plantations' trees but other trees may be visible in the imagery -- for example, trees outside the plantation area on the border of the orthomosaics. We manually delineated areas of interest (AOIs) on QGIS to exclude trees that do not have a corresponding annotation in the imagery. 
We use the \textit{geodataset}\footnote{\url{https://hugobaudchon.github.io/geodataset/index.html}}  Python package to process the AOIs into 50\% overlapping $1024\times1024$ pixels tiles. We exclude tiles which contain no label, and tiles with more than 80\% black pixels at the border of AOIs. We also exclude annotations when less than 20\% of the tree appears in a given tile. We align the DSMs to the resolution of the RGB orthomosaics. 
We split the data into training, validation and test sets, defining  polygonal regions corresponding to geographical blocks to avoid spatial autocorrelation, and ensure that each class is represented in all sets. Orthomosaics are either assigned entirely to a split, or assigned to different splits by manually delineating areas on QGIS. Our final image dataset is composed 26,428 tiles and the training, validation and test sets contain respectively 15,742, 6,691 and 3,995 tiles. Table \ref{species_dist_per_split} in \ref{appendix:data} summarizes the number of annotations per tree species per split.

\section{Methods}\label{method}
We investigate both the relevance of SAM and the informativeness of the DSM for tree crown instance segmentation. 
We compare different methods, with and without different components of SAM, and for each we propose a version where the DSM is used as input along with the RGB imagery.
We hypothesize that the DSM is a useful input, helping capture the vertical structure of sites, especially since the trees considered in this study are from 10-year-old plantations and relatively small, and the canopy is more open compared to trees from more widely available forest datasets. We detail choices of backbones and hyperparameters in Appendix \ref{appendix:setup}. \\

\textbf{SAM out-of-the-box.} We first assess to what extent SAM can segment tree crowns in our dataset without additional training or tuning. We benchmark SAM in the automatic mask generation mode (denoted \textit{SAM}) and SAM using point prompts defined as the local elevation maxima in the DSM in each image (denoted \textit{SAM+DSM prompts}). An overview of SAM+DSM prompts is shown in Figure \ref{fig:dsmprompts} in Appendix \ref{appendix:models}. For both models we apply Non-Maximum Suppression (NMS) on the segmented instances. We also considered using the DSM image as a dense prompt but obtained very poor segmentation masks, as dense prompts are intended to be binary masks (see Appendix \ref{appendix:samdsmmask}).\\
\textbf{Mask R-CNN and variations.} We consider Mask R-CNN as a comparison as this architecture has previously been used successfully for semantic segmentation of aerial imagery into trees/non-trees pixels \citep{ball2023detectree}. We compare both a Mask-RCNN initialized with weights from a model pre-trained on ImageNet and trained from scratch. We also train a model taking as input the DSM as a fourth channel, stacking it onto the RGB image (\textit{Mask R-CNN$+$DSM}).\\
\textbf{Faster/Mask-CNN$+$SAM and variations.} 
In these methods, we trained a Faster-RCNN for the task of object detection on our dataset. We then fed the predicted detections as box prompts to SAM. We also try the same with Mask-RCNN, feeding both the predicted boxes and  segmentation masks as prompts (respectively box and mask) to SAM. 
Similarly to the Mask R-CNN baseline, we also include the DSM as input by stacking it to the RGB images as a fourth channel for both Faster-RCNN$+$SAM and Mask-CNN$+$SAM. 
\begin{wrapfigure}{r}{0.5\textwidth}
    \centering   \includegraphics[width=1.1\linewidth]{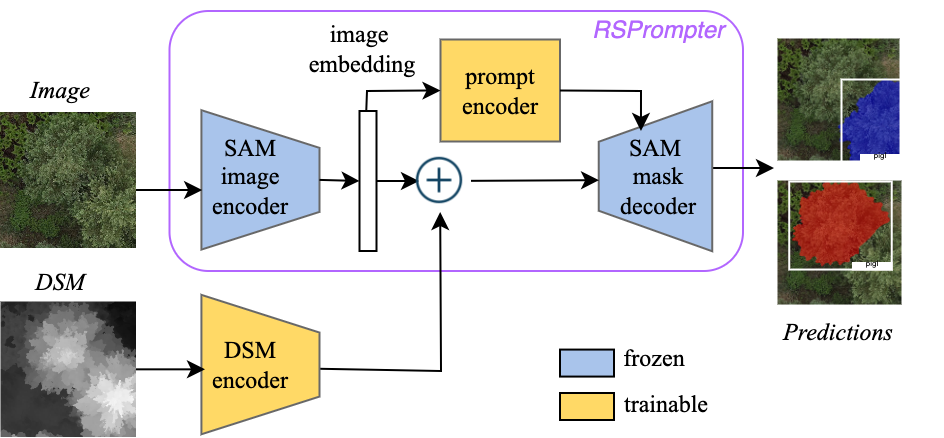}
    \caption{\small Overview of our DSMPrompter method.}
    \label{fig:overview}
\end{wrapfigure}\\
\textbf{RSPrompter and variations.} We include RSPrompter in our study as the method was originally developed specifically for remote sensing applications, and we train the RSPrompter-anchor version of it following the original paper. 
We also propose to enhance RSPrompter by integrating the DSM as input (\textit{DSMPrompter}), as shown in Fig.~\ref{fig:overview}. We add a trainable DSM encoder to RSPrompter, following the same architecture as the SAM dense prompt encoder. The encoded DSM is added to the image embedding before being fed as input to the mask decoder.  

\subsection{Evaluation}

We evaluate our instance segmentation models with mean Average Precision (mAP). Given the class imbalance in our dataset, we also consider a weighted mAP (wmAP) by the number of examples in each class.
Since SAM out-of-the-box methods provide segmentation masks of each instance but no associated class
label, we also evaluate the models with mAP considering the single class
``trees'' to compare methods.
Finally, we also consider mean Intersection over Union (mIoU) considering the single class ``trees":  for each ground truth instance, we take
the instance in the prediction for which IoU is highest and average this IoU  over all instances in the dataset. Note that this metric doesn’t reflect false positive instances, as it only compares each ground truth instance with a single predicted instance -- namely, the best matching one in terms of IoU. We can consider that this metric reflects the quality of the segmentation if the object has been correctly detected. 

\section{Results}

Table \ref{results} summarizes the models' performance in terms of single-class ``tree" metrics and aggregated mAP metrics over the classes. Per class mAP performance is reported in Table \ref{mapperclass} in Appendix \ref{appendix:results}. For the models which can be evaluated with classwise mAP, we prioritize wmAP to assess the performance of the models, as our dataset has very unbalanced classes. 

 Our first observation is that SAM out-of-the-box methods perform poorly on our task. Qualitatively, we observe that in many cases, SAM automatic fails to separate touching crowns into separate masks and confidently segments the background or tiny plants, leading to many false positives. It also misses trees in areas where tall herbaceous vegetation occurs. Examples of segmentations are provided in Figure~\ref{appendix:sam_failure}. SAM$+$DSM prompts predictions also exhibit many false positives as each local maxima detected in the DSM will correspond to a segmented object. Hence local maxima corresponding to tall grass, or small plants standing out on the ground might be given as prompts to the mask decoder as shown in Figure \ref{fig:exampleprompts} in Appendix~\ref{appendix:models}. However, when prompts corresponding to a treetop are given, SAM is able to correctly segment the tree crown, explaining the boost in mIoU compared to SAM automatic. \\
 Mask R-CNN methods perform well, and initialization of the ResNet-50 backbone with ImageNet weights improves performance, compared to training from scratch. Adding the DSM as input improves further the performance, on all metrics. \\

Surprisingly, we find that using boxes and mask outputs of the Mask R-CNN models as prompts to SAM degrades performance compared to the Mask R-CNN model. In fact, when looking at predictions, we observe that SAM sometimes focuses on very small details and artifacts in the imagery, degrading the quality of the original segmentation. Examples of segmentations are shown in Figure \ref{appendix:samsegmaskrcnnprompt} in Appendix \ref{appendix:results}.Similarly, we find that Faster R-CNN+SAM models perform significantly worse than Mask R-CNN, and this drop is heavily attributable to the performance on the \textit{Picea glauca} and \textit{Thuya occidentalis} classes. However, adding DSM consistently improves performance for all models. 
\begin{wrapfigure}{}{0.45\textwidth} 
    \centering
    \includegraphics[width=0.45\textwidth]{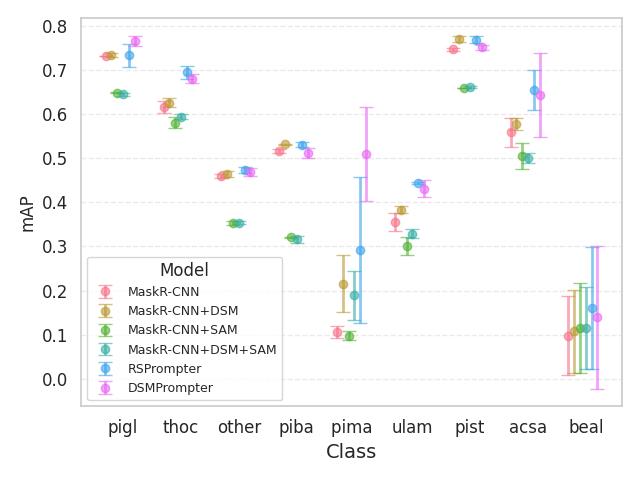} 
    \caption{\footnotesize Per class mAP performance on the test set. For each model, the performance is averaged on 3 seeds. Tree species on the x-axis are ordered by decreasing prevalence in the dataset from left to right. Mask R-CNN is pretrained on ImageNet. Numerical results are provided in Appendix \ref{appendix:results}.}
    \label{fig:mapperclass}
    \vspace{-\baselineskip}
\end{wrapfigure}

RSPrompter performs significantly better than Mask R-CNN, especially in terms of wmAP, the most relevant metric for our application. DSMPrompter achieves the best performance, though its performance is comparable to  RSPrompter, and integrating the DSM information does not provide a significant advantage to RSPrompter. But, we note that for the DSMPrompter models, the best model (on the validation set) is obtained after training for 7 epochs on average compared to 20 for RSPrompter. Figure \ref{fig:dsm-rsprompter} in Appendix \ref{dsmrsappendix} shows example predictions.\\
 When looking closer into the per class performance for the instance segmentation models (i.e. excluding SAM out-of-the-box based methods), we observe a general pattern of performance increasing with the number of examples for a given class for all models, as shown in Figure \ref{fig:mapperclass}. However, despite there being few examples of this class, all methods perform relatively well on \textit{Acer saccharum}  (acsa) which looks quite visually different from the rest of the species in the dataset.  The performance of the different models differs most on the \textit{Pinus mariana} (pima) class. In fact, it is a very similar species to the most common class in our dataset \textit{Pinus strobus} (pist) and one that is identified most easily by looking at the shapes of the cones on the tree when conducting field surveys rather than other characteristics.
 
\begin{table}[]
\resizebox{1\textwidth}{!}{
\centering
\begin{tabular}{lcc| ll | ll }
      & & & \multicolumn{2}{c|}{Single-class} & \multicolumn{2}{c}{Multi-class} \\ \hline\hline
Model & DSM & Pre-trained  & mAP             & mIoU            & mAP            & wmAP            \\ \hline\hline
 SAM  (100 pps)  & \xmark & --    &        8.05         &  35.06               &        \multicolumn{1}{c}{--}     &      \multicolumn{1}{c}{--}      \\
  SAM  (10 pps)   & \xmark &  --  &        10.11         &      34.01           &         \multicolumn{1}{c}{--}       &      \multicolumn{1}{c}{--}              \\
 SAM & \cmark (prompts) & --  &        9.28         &      46.15           &          \multicolumn{1}{c}{--}      &     \multicolumn{1}{c}{--}              \\ \hline
 \multirow{3}{*}{Mask R-CNN}   & \xmark & \xmark    &     59.36 \scriptsize$(\pm0.21$)            &    79.63 \scriptsize $(\pm0.53$)             &     42.69 \scriptsize$(\pm2.93$)           &      55.75 \scriptsize$(\pm1.50$)           \\
                                & \xmark & \cmark  &  63.65 \scriptsize $(\pm    0.43)$       &    81.82 \scriptsize $( \pm 0.36)$             &        46.51 \scriptsize $(\pm1.13)$      &     58.30 \scriptsize $(\pm1.23$)        \\
                                & \cmark &  \cmark    &       64.64 \scriptsize $(\pm0.69$)          &   81.89 \scriptsize $(\pm0.60$)              &       48.96 \scriptsize $(\pm1.06$)         &          60.32 \scriptsize $(\pm0.73$)     \\ \hline
 \multirow{3}{*}{Faster R-CNN$+$SAM}     & \xmark &  \xmark   &   53.56 \scriptsize($\pm0.20)$   &  76.22 \scriptsize $(\pm0.20)$  &  33.52 \scriptsize$(\pm0.44)$ &    45.79 \scriptsize $(\pm0.67)$
    \\ 
                                        & \xmark & \cmark   &     57.85 \scriptsize($\pm0.66)$  &   78.0 \scriptsize($\pm0.56)$  &  39.79 \scriptsize($\pm1.17)$ &   50.30 \scriptsize $(\pm1.51)$            \\ 
                                         & \cmark & \cmark    &    58.0 \scriptsize($\pm0.25)$  & 78.27 \scriptsize $(\pm0.74)$  &    40.14  \scriptsize $(\pm1.41)$  &  52.08 \scriptsize $(\pm1.73)$                 \\ \hline
 \multirow{2}{*}{Mask R-CNN$+$SAM}     & \xmark &  \cmark   & 57.6 \scriptsize $(\pm0.19)$  & 78.18 \scriptsize $(\pm0.31)$  &  39.76  \scriptsize $(\pm1.19)$            &     50.46 \scriptsize $(\pm 0.52$)       \\ 
                                         & \cmark & \cmark   &  57.83 \scriptsize $(\pm0.11)$  & 77.65 \scriptsize $(\pm0.5)$ &  41.13 \scriptsize $(\pm1.12)$ &   51.33 \scriptsize $(\pm
0.85$)         \\ \hline
RSPrompter      & \xmark & --   &      66.37 \scriptsize $(\pm0.91$)           &   82.58 \scriptsize $(\pm1.63$)             &   52.77 \scriptsize $(\pm1.03$)             &       62.37  \scriptsize $(\pm2.45$)            \\ 
DSMPrompter      & \cmark &  --  &     65.03 \scriptsize $(\pm1.76$)            &            83.24 \scriptsize $(\pm0.41$)     &   54.40 \scriptsize $(\pm4.00$)             &    64.84 \scriptsize $ (\pm1.49$)  

\end{tabular}
 }
\caption{\small Results on the test dataset, averaged over 3 seeds. All metrics are multiplied by $10^2$. The column \textit{Pre-trained} refers to ImageNet pretraining for the backbones of the Mask R-CNN and Faster R-CNN  models (SAM is always pretrained); ``--'' denotes not applicable.}
\label{results}
\end{table}

\section{Conclusion and future work}
In this work, we conducted the first study of tree crown instance segmentation in high-resolution imagery on the UAV Quebec Plantations dataset. We explored different ways in which SAM can be leveraged and assessed the relevance of the use of this foundational model for our task. We find that methods using SAM for inference  without further tuning do not outperform a simpler Mask R-CNN model trained specifically for our task of interest. Our study highlights the limitations of SAM, despite its presentation as an out-of-the-box tool, in the context of automating tree plantations monitoring. However, when tuned properly, SAM components can be leveraged to make it a powerful tool, as in RSPrompter. 
We also show that including DSM information improves predictions of all models. Promising future directions include studying different ways for integrating the DSM into models, assessing the effect of downgrading the resolution of the input imagery, and determining whether DSMPrompter could offer an even greater advantage over Mask R-CNN-based methods in the regime of low training data.

\bibliography{iclr2025_conference}
\bibliographystyle{iclr2025_conference}

\newpage
\appendix
\section{Appendix}
\subsection{Dataset}\label{appendix:data}
In this section, we provide more details about the composition of the dataset and the splits that we used in this study. 
\begin{table}[h] 
\centering
\begin{tabular}{l|p{1.5in}}
piba  & \textit{Pinus banksiana}                                                                     \\
pima  & \textit{Picea mariana}                                                                       \\
pist  & \textit{Pinus strobus}                                                                       \\
pigl & \textit{Picea glauca}                                                                               \\
thoc  & \textit{Thuya occidentalis}                                                                  \\
ulam  & \textit{Ulmus americana}                                                                     \\
beal  & \textit{Betula allegnaniensis}                                                               \\
acsa  & \textit{Acer Saccharum}                                                                      \\
other & Other, \textit{Larix laricina, Pinus resinosa, Populus tremuloides, Betula papyrifera, Quercus rubra}
\end{tabular}
\caption{Species code and scientific name correspondence}
\label{specie_name_correspondence}
\end{table}

\begin{table}[h]\label{num_trees_per_site}
\begin{tabular}{l|lllllllll|l}
                                  & \multicolumn{1}{l}{piba}  & \multicolumn{1}{l}{pima}    & \multicolumn{1}{l}{pist}    & \multicolumn{1}{l}{pigl}    & \multicolumn{1}{l}{thoc}    & \multicolumn{1}{l}{ulam}    & \multicolumn{1}{l}{beal}   & \multicolumn{1}{l}{acsa}  & \multicolumn{1}{l}{other}                      & \multicolumn{1}{l}{total}    \\
                                  \hline
 cbpapinas &  0 &  136 &  121 &  1437 &  182 &  142 &  11 &  5 & 32& 2076     \\ 
cbblackburn1                      & 1440                      & 215                         & 102                         & 100                          & 0                           & 0                           & 1                          & 0                         & 7                                              &  1865 \\
 
cbblackburn2                      & 573                       & 18                          & 50                          & 993                          & 0                           & 1                           & 0                          & 0                         & 67                                             &  1702 \\

cbblackburn3                      & 0                         & 0                           & 0                           & 140                          & 3                           & 0                           & 0                          & 0                         & 2                                              &  145  \\
cbblackburn4                      & 278                       & 0                           & 0                           & 11                           & 355                         & 125                         & 0                          & 0                         & 6                                              &  775  \\
cbblackburn5                      & 86                        & 0                           & 0                           & 514                          & 0                           & 0                           & 0                          & 0                         & 3                                              &  603  \\
cbblackburn6                      & 3002                      & 273                         & 122                         & 1746                         & 216                         & 149                         & 2                          & 0                         & 3                                              &  5513 \\
cbbernard1                        & 0                         & 0                           & 0                           & 221                          & 0                           & 0                           & 0                          & 0                         & 14                                             &  235  \\
cbbernard2                        & 0                         & 0                           & 14                          & 61                           & 0                           & 0                           & 0                          & 0                         & 0                                              &  75   \\ 
cbbernard3                        & 0                         & 283                         & 377                         & 531                          & 7                           & 2                           & 8                          & 73                        & 19                                             &  1300 \\
cbbernard4                        &  0 &  0   &  206 &  1193 &  0   &  0   &  1  &  0 &  2                      & 1402 \\
afcamoisan                        & 0                         & 0                           & 0                           & 628                          & 0                           & 0                           & 0                          & 0                         &  2                      & 630                          \\
afcahoule                         & 0                         & 0                           & 0                           & 1004                         & 0                           & 0                           & 0                          & 0                         & 1                      & 1005                         \\
afcagauthmelpin                   & 0                         & 0                           & 0                           & 0                            & 0                           & 0                           & 0                          & 0                         & 1674                   & 1674                             \\
afcagauthier                      & 0                         & 0                           & 0                           & 500                          & 0                           & 0                           & 0                          & 0                         & 0                      & 500    \\ \hline\hline
Total &5379	&925&	992&	9079	&763	&419&	23&	78&	1842	& 19500
\end{tabular}
\caption{Number or trees per species per site}
\end{table}

\begin{table}[h]
\centering
 \begin{tabular}{l|l|l|l|l|l|l|l|l|l|l } 
      & piba  & pima & pist & pigl & thoc & ulam & acsa & beal & other & \textbf{total} \\ \hline
train & 19869 & 1377 & 2224 & 32496 & 1079 & 709  & 179  & 51   & 3343  & 61327          \\ \hline
val   & 6978  & 2046 & 2447 & 6710  & 573  & 245  & 116  & 40   & 3713  & 22868          \\\hline
test  & 1471  & 1056 & 544  & 6519  & 1946 & 1050 & 56   & 19   & 1601  & 14262         
\end{tabular}
\caption{Tree species annotations distribution in the different splits. Note the values for each set and species are higher than the number of trees because tiles have 50\% overlap.}
\label{species_dist_per_split}
\end{table}

\section{Models}\label{appendix:models}

\subsection{SAM+DSM prompts}
\begin{figure}
    \centering  
    \includegraphics[width=0.5\linewidth]{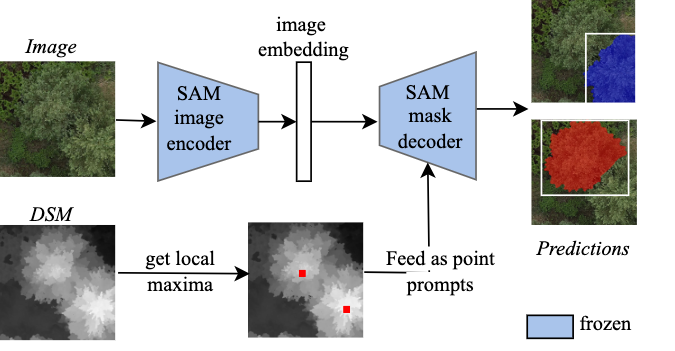}
    \caption{Overview of our SAM$+$DSM prompts method.}
    \label{fig:dsmprompts}
\end{figure}

In Figure \ref{fig:dsmprompts}, we show an overview of the SAM$+$DSM prompts method described in Section \ref{method}.

\begin{figure}
    \centering  
    \includegraphics[width=0.5\linewidth]{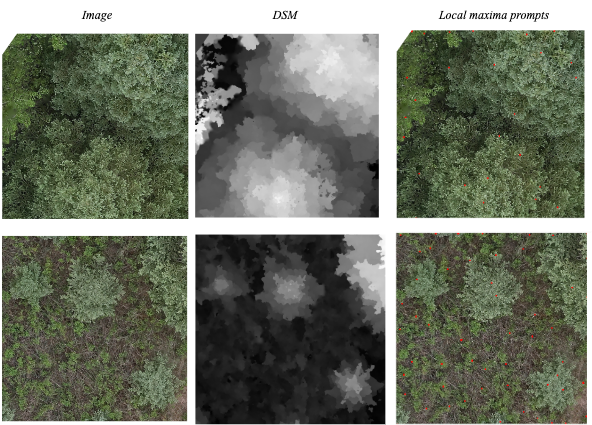}
    \caption{Examples of image, DSM and local maxima prompts (red dots) overlayed on the image. }
    \label{fig:exampleprompts}
\end{figure}

Figure \ref{fig:exampleprompts} shows some examples of local maxima that are fed as prompts to the mask decoder. One limitation of this method is in the case where there are a lot of small plants sticking out of the ground, giving many local maxima prompts that do not correspond to a tree. Note that if multiple prompts are given on a tree, it should not be too much of a limitation since we apply NMS to the predictions. 

\subsection{SAM+DSM mask prompts}\label{appendix:samdsmmask}
\begin{figure}
    \centering  
    \includegraphics[width=0.35\linewidth]{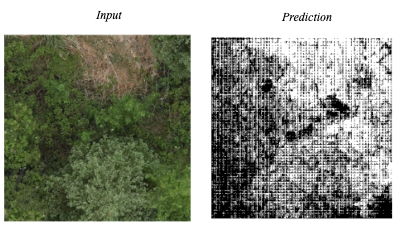}
    \caption{Examples of image and prediction when the DSM is fed as a mask prompt to SAM.}
    \label{fig:failuredsmprompt}
\end{figure}

We also tried feeding a normalized DSM to SAM as a mask prompt. SAM normally calls for binary mask prompts, and feeding the DSM as a mask prompt would give gridded segmentations which were not satisfactory enough to be included in this study, as shown in Figure \ref{fig:failuredsmprompt}.
\section{Setup}\label{appendix:setup}

In all experiments, we use the ViT-Huge version of SAM. 
We first evaluate SAM in its automatic mode on the test set tiles with a points per side (pps) value of 100 (default parameter) and 10.
For SAM+DSM prompts, the local maxima in the DSM are obtained using \verb|scipy.ndimage.maximum_filter| with a size parameter of 100. For all SAM out-of-the-box methods, NMS is applied on the predictions with a score threshold of 0.5 and overlap IoU threshold of 0.5. \\
All Mask R-CNN-based models use the \verb|torchvision| implementation of Mask R-CNN are trained for a maximum of 100 epochs, and SGD optimizer with learning rate 0.0001, momentum 0.9, and weight decay 0.0005, and linear warmup starting at $10^{-6}$. Batch size is 32 for Mask R-CNN and 8 for Mask R-CNN$+$DSM. NMS is applied with the default parameters. We initialize the ResNet-50 backbone of MaskRCNN$+$DSM with ImageNet weights, and for the first layer, copy the weights to the channels corresponding to the RGB input.\\

All Faster R-CNN-based models use the \verb|torchvision| implementation of Faster R-CNN are trained for a maximum of 100 epochs, and Adam optimizer with learning rate 0.0001 for finetuning and 0.0005 when trained from scratch, betas of 0.9 and 0.999, weight decay of 0.0005, and using an exponential decay scheduler updating the learning rate each 10 epochs. Batch size is 32 for Faster R-CNN and 16 for Faster R-CNN$+$DSM. NMS is applied with the default parameters. We initialize the ResNet-50 backbone of Faster R-CNN$+$DSM with ImageNet weights, and for the first layer, copy the weights to the channels corresponding to the RGB input.\\

For Faster R-CNN$+$SAM and Mask R-CNN$+$SAM methods, the scores used to compute the mAP metrics are the average of the output scores of Faster R-CNN / Mask R-CNN and SAM predicted IoU scores. We also considered using only Faster R-CNN / Mask R-CNN scores but did not observe that it made a significant difference. 

Following \citet{chen2024rsprompter}, 
the RSPrompter based methods are trained with input images of size 1024$\times$1024, normalized with ImageNet statistics, and learning rate scheduler strategy of linear warmup followed by cosine annealing. The models are trained with batch size 2, base learning rate of 0.00001 with linear warmup starting at $10^{-8}$ for one epoch followed by cosine annealing. We use AdamW optimizer with weight decay 0.1.\\
For all trained models, we apply RandomFlip augmentations during training and normalize the DSM per sample by its maximum value.
We select the best model based on the validation segmentation mAP value (over all classes). 

\section{Results}\label{appendix:results}
\subsection{mAP per class}
In Table \ref{mapperclass}, we report the per class mAP on the test set for the different methods in our study, to the exception of the SAM out-of-the-box methods which do not classify the predicted masks.
\begin{table}[h]
\resizebox{1.2\textwidth}{!}{
    \hspace{-3cm}
    \centering
    \begin{tabular}{lcc|lllllllll}
Model   &DSM       &Pretrained                  & piba   & pima   & pist   & pigl   & thoc   & ulam   & other  & beal   & acsa   \\ \hline\hline

Mask R-CNN   &\xmark     &\xmark         & 45.88\scriptsize ($\pm$0.14) & 13.31 \scriptsize ($\pm$10.17)&73.60 \scriptsize ($\pm$0.52)& 69.75 \scriptsize ($\pm$1.10)& 60.08 \scriptsize ($\pm$0.74)& 30.90 \scriptsize ($\pm$2.34)& 41.82 \scriptsize ($\pm$0.21)& 7.64 \scriptsize ($\pm$11.77)& 41.23  \scriptsize ($\pm$6.35)\\
Mask R-CNN &\xmark &  \cmark      & 51.55 \scriptsize ($\pm$0.50)& 10.59\scriptsize ($\pm$1.30) & 74.67 \scriptsize ($\pm$0.35)& 73.14 \scriptsize ($\pm$0.03)& 61.53 \scriptsize ($\pm$1.34)& 35.49 \scriptsize ($\pm$1.99)& 45.94 \scriptsize ($\pm$0.48)& 9.81 \scriptsize ($\pm$8.98)& 55.78 \scriptsize ($\pm$3.25)\\

Mask R-CNN  &\cmark    &  \cmark  & 53.08 \scriptsize ($\pm$0.17) & 21.56\scriptsize ($\pm$6.38) & 76.97\scriptsize ($\pm$0.74) & 73.29 \scriptsize ($\pm$0.50)& 62.57 \scriptsize ($\pm$1.05)& 38.30 \scriptsize ($\pm$0.74)& 46.43 \scriptsize ($\pm$0.70) & 10.78 ($\pm$9.39)& 57.69 \scriptsize ($\pm$1.32)\\
Faster R-CNN+SAM &\xmark  &\xmark  &60.56	\scriptsize ($\pm$0.09)&56.28	\scriptsize ($\pm$0.88)&28.96\scriptsize ($\pm$0.91)	&24.53	\scriptsize ($\pm$1.41)&3.32	\scriptsize ($\pm$0.53)&26.26	\scriptsize ($\pm$1.96)&60.51	\scriptsize ($\pm$1.45)&41.2\scriptsize ($\pm$6.87)&	0 \scriptsize ($\pm$0.0)\\
Faster R-CNN+SAM &\xmark  &\cmark  &64.11	\scriptsize ($\pm$1.13)&61.03	\scriptsize ($\pm$1.39)&34.93	\scriptsize ($\pm$1.02)&31.57	\scriptsize ($\pm$0.64)&3.61	\scriptsize ($\pm$3.95)&33.58	\scriptsize ($\pm$4.11)&66.65 \scriptsize ($\pm$0.18)&	49.82	\scriptsize ($\pm$3.98)&12.84 \scriptsize ($\pm$5.04)\\
Faster R-CNN+SAM &\cmark  &\cmark   & 66.04	\scriptsize ($\pm$1.18)& 61.38	\scriptsize ($\pm$0.47)& 35.46	\scriptsize ($\pm$0.1)& 30.78	\scriptsize ($\pm$0.32)& 17.26	\scriptsize ($\pm$13.61)& 31.94 \scriptsize ($\pm$0.68)& 	66.37 \scriptsize ($\pm$0.61)& 	51.86 \scriptsize ($\pm$1.5)& 	0.15 \scriptsize ($\pm$0.26)\\
RSPrompter    &\xmark     &  --              & 53.03 \scriptsize ($\pm$0.51)& 29.17 \scriptsize ($\pm$16.50)& 76.83 \scriptsize ($\pm$0.81)& 73.23 \scriptsize ($\pm$2.51)& 69.43 \scriptsize ($\pm$1.48)& 44.40 \scriptsize ($\pm$0.20)& 47.33 \scriptsize ($\pm$0.64)& 16.00 \scriptsize ($\pm$13.84)& 65.43 \scriptsize ($\pm$4.59)\\
DSMPrompter  &\cmark    &  --              & 51.17 \scriptsize ($\pm$1.19)& 50.97 \scriptsize ($\pm$10.65)& 75.07 \scriptsize ($\pm$0.60)& 76.47 \scriptsize ($\pm$1.06)& 67.93 \scriptsize ($\pm$0.99)& 43.07 \scriptsize ($\pm$1.94)& 46.87 \scriptsize ($\pm$0.93) & 13.93 \scriptsize ($\pm$16.14) & 64.23  \scriptsize ($\pm$9.58)\\

    \end{tabular}
    }
    \caption{mAP per class [$10^2$] on the test set for the instance segmentation models in our study.}
    \label{mapperclass}
\end{table}
\subsection{Qualitative results}
\subsubsection{SAM automatic outputs}
\begin{figure}[h]
    \centering
    \includegraphics[width=\linewidth]{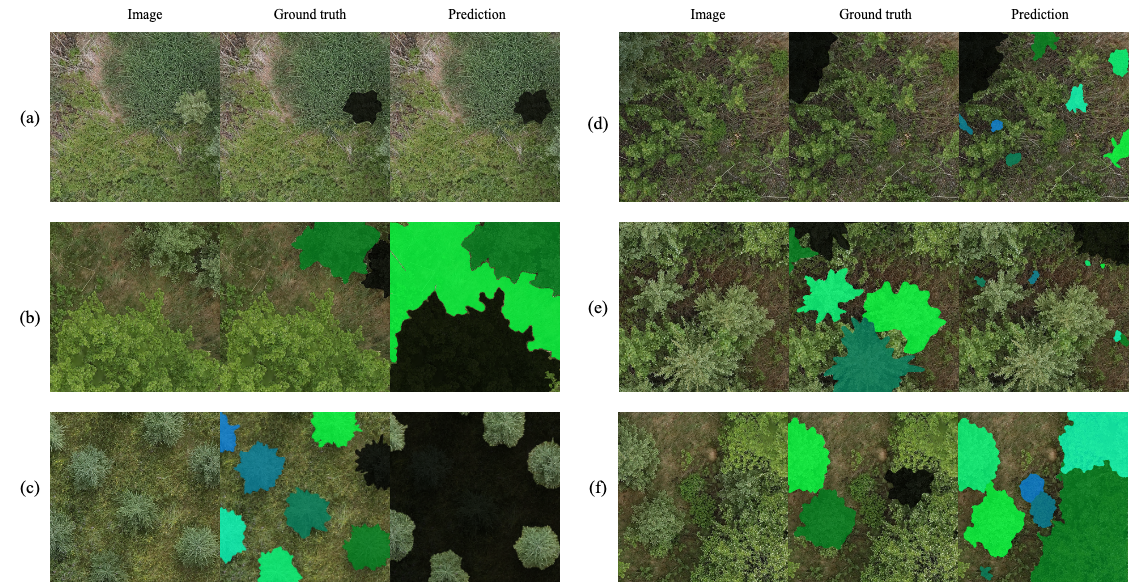}
    \caption{Examples of SAM automatic predictions: (a) A success case. (b) SAM segments everything, including the background, and merging two touching crowns into a single instance in the top right corner. (c) SAM segments only the background, i.e., everything but the objects of interest. (d) A lot of tiny isolated objects are segmented. (e) SAM completely misses the objects of interest. (f) SAM segments the trees and also the large bushes around. }
    \label{appendix:sam_failure}    
\end{figure}

Figure \ref{appendix:sam_failure} shows example predictions of SAM when it is used in its automatic mode.
\subsubsection{Mask R-CNN vs. Mask R-CNN+SAM}
\begin{figure}[h]
    \centering
    \includegraphics[width=0.9\linewidth]{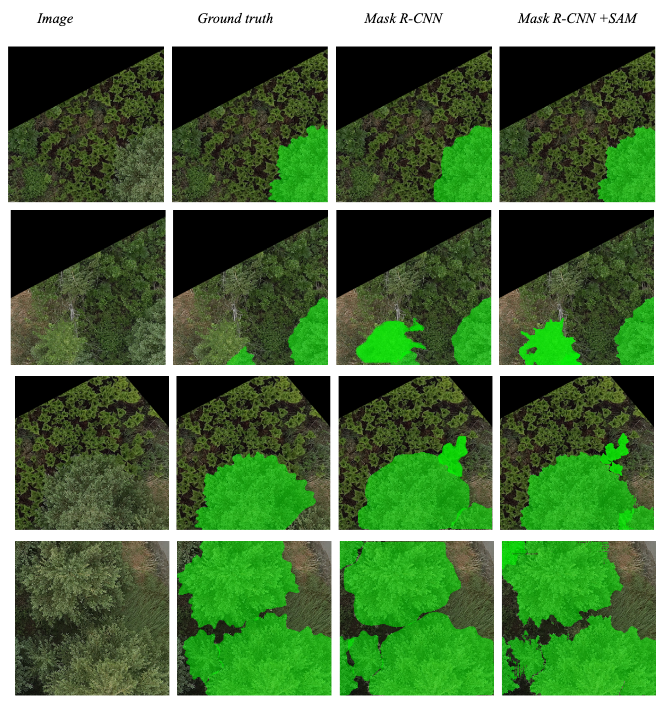}
    \caption{Comparison of segmentations obtained with Mask R-CNN and Mask R-CNN outputs fed as prompts to SAM. On the last row, the quality of the mask is degraded when outputs of Mask R-CNN are passed as prompts to SAM, compared to the outputs of Mask R-CNN. In Figure \ref{fig:fullsize}, we can see that locations where the masks look "gridded" correspond to artifacts in the input image that come from constructing the original orthomosaic with photogrammetry.}
    \label{appendix:samsegmaskrcnnprompt}
\end{figure}

\begin{figure}
    \centering
    \includegraphics[width=0.9\linewidth]{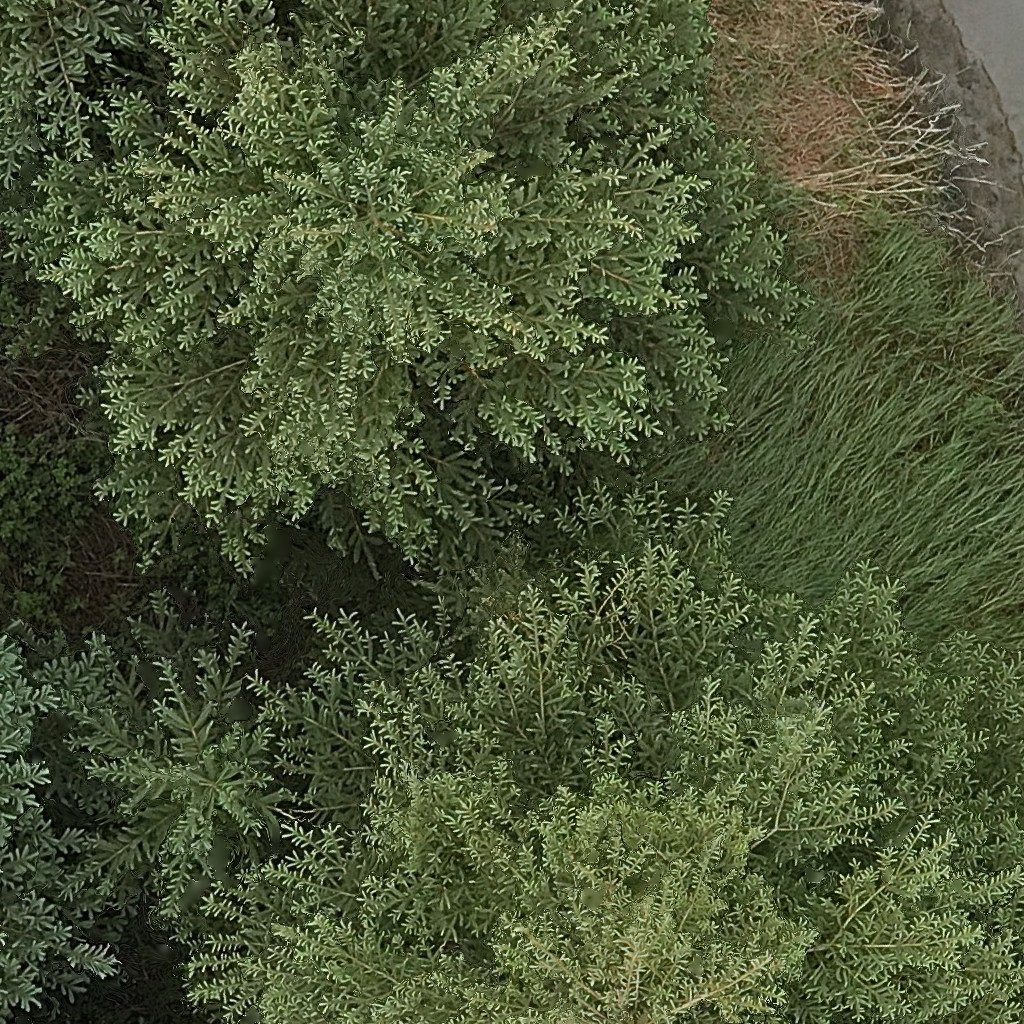}
    \caption{Sample tile of our dataset.}
    \label{fig:fullsize}
\end{figure}

Figure \ref{appendix:samsegmaskrcnnprompt} compares predictions from Mask R-CNN, and from SAM when predicted boxes and masks from Mask R-CNN are passed as prompts. While in some cases, SAM does indeed refine Mask R-CNN masks (first row), the high resolution and the artifacts (from the RGB orthomosaic processing) in the imagery can lead to gridded segmentations (bottom row). We show a the full resolution image of the bootom row in Figure \ref{fig:fullsize} where if you zoom in, you can see artefacts due to the process of obtaining orthomosaics from drone imagery.

\subsection{RSPrompter and DSMPrompter predictions}\label{dsmrsappendix}
\begin{figure}
    \centering
    \includegraphics[width=0.9\linewidth]{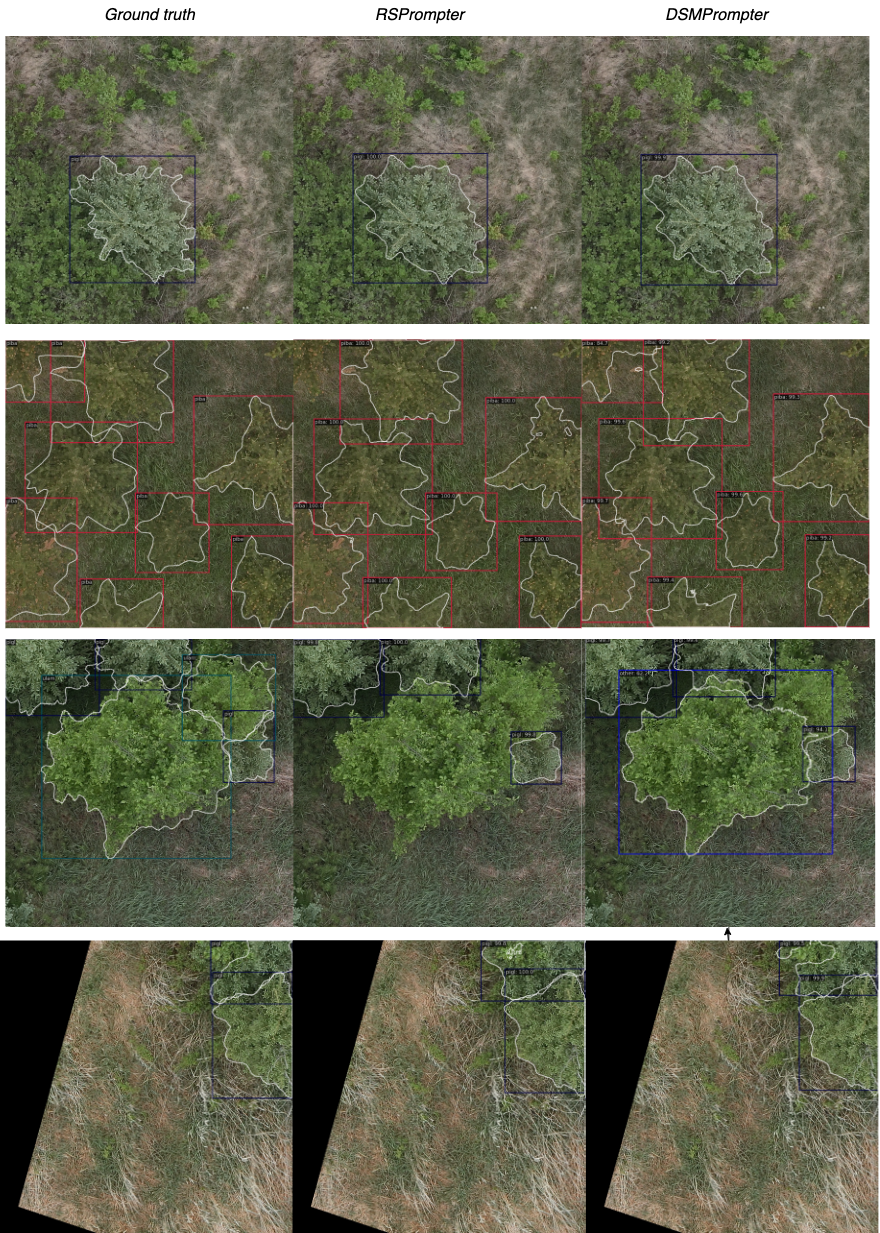}
    \caption{Examples of ground truth, RSPrompter and DSMPrompter predictions. Generally, DSMPrompter segmentations are sharper.}
    \label{fig:dsm-rsprompter}
\end{figure}

In Figure \ref{fig:dsm-rsprompter}, we show example predictions of RSPrompter and DSMPrompter. 

\end{document}